\documentclass[final, runningheads]{llncs}
\usepackage[dvipsnames]{xcolor}
\usepackage[T1]{fontenc}
\usepackage{graphicx}
\usepackage{subcaption} \usepackage{caption}
\usepackage{array}

\usepackage{amsmath}
\usepackage{bm}

\usepackage{enumerate}

\usepackage{todonotes}

\usepackage{hyperref}

\hypersetup{
    colorlinks=true,
    linkcolor=blue,
    filecolor=magenta,      
    urlcolor=cyan,
    pdftitle={Overleaf Example},
    pdfpagemode=FullScreen,
    citecolor=black,
    }
\newcommand{\itadata}{\footnotesize \textsl{ITADATA2024: The 3$^{\text{rd}}$ Italian Conference on Big Data and Data Science}}
\usepackage{fancyhdr}
\fancypagestyle{empty}{\fancyhf{}\fancyhead[C]{\itadata}
  \fancyhead[R]{\includegraphics[width=.05\textwidth]{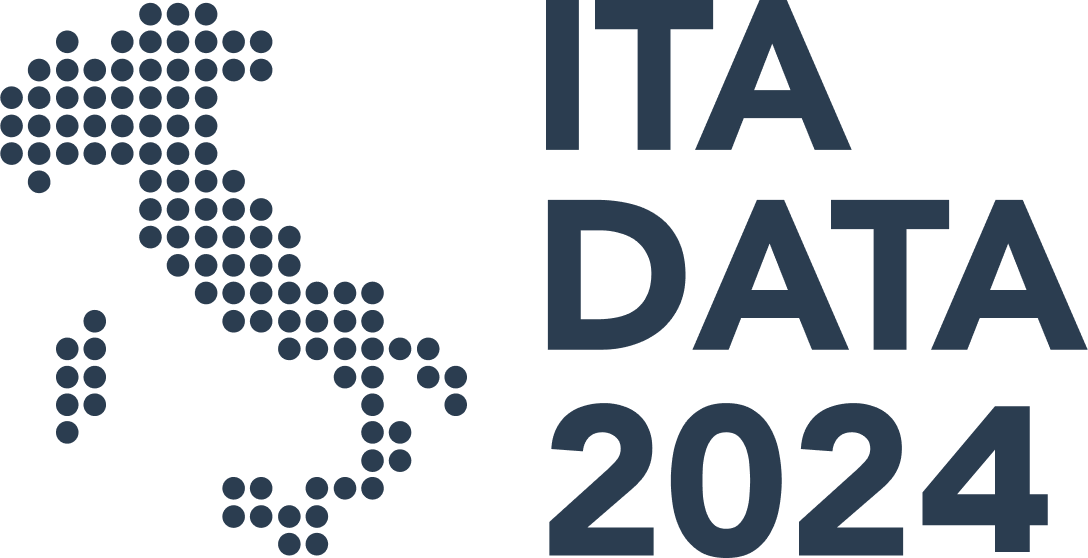}}}

\makeatletter
\begin{document}
\title{Multivariate Time Series Anomaly Detection in Industry 5.0}
\author{Lorenzo Colombi\inst{1}\orcidID{0009-0004-9681-9842} \and
Michela Vespa \inst{1}\orcidID{0009-0004-4350-8151} \and
Nicolas Belletti\inst{1}\orcidID{0009-0006-6053-433X} \and
Matteo Brina\inst{1} \orcidID{0009-0003-2363-8509}\and
Simon Dahdal \inst{1} \orcidID{0000-0002-7335-9773} \and
Filippo Tabanelli\inst{1} \orcidID{0009-0008-9661-394X} \and
Elena Bellodi\inst{1} \orcidID{0000-0002-3717-3779} \and
Mauro Tortonesi\inst{1} \orcidID{0000-0002-7417-4455} \and
Cesare Stefanelli\inst{1} \orcidID{0000-0003-4617-1836} \and
Massimiliano Vignoli \inst{2}
}
\authorrunning{L. Colombi et al.}
\institute{University of Ferrara, Ferrara, Italy 
\email{
\{firstname.lastname\}@unife.it}\and
Bonfiglioli S.P.A., Via Cav. Clementino Bonfiglioli, 1, Calderara di Reno, Italy
\email{massimiliano.vignoli@bonfiglioli.com}
\vspace{-20pt}
}
\maketitle \begin{abstract}
Industry 5.0 environments present a critical need for effective anomaly detection methods that can indicate equipment malfunctions, process inefficiencies, or potential safety hazards.
The ever-increasing sensorization of manufacturing lines makes processes more observable, but also poses the challenge of continuously analyzing vast amounts of multivariate time series data. These challenges include data quality since data may contain noise, be unlabeled or even mislabeled. 
A promising approach consists of combining an embedding model with other Machine Learning algorithms to enhance the overall performance in detecting anomalies. Moreover, representing time series as vectors brings many advantages like higher flexibility and improved ability to capture complex temporal dependencies. 
We tested our solution in a real industrial use case, using data collected from a Bonfiglioli plant. The results demonstrate that, unlike traditional reconstruction-based autoencoders, which often struggle in the presence of sporadic noise, our embedding-based framework maintains high performance across various noise conditions.

\end{abstract}
\section{Introduction}

In digital manufacturing, innovations in Machine Learning (ML) enable to boost productivity and gain competitive advantages by predicting equipment failures and improving supply chain efficiency through real-time forecasting. ML and MLOps applications optimize industrial processes by using data-driven insights to refine production methods, enhance product quality, and increase operational efficiency \cite{brinacolombi}. Research is advancing towards the Industry 5.0 paradigm, and its challenging Zero Defect Manufacturing (ZDM) and Zero Waste Manufacturing (ZWM) objectives, aiming for more sustainable processes \cite{colombi2024mlops,VENANZI2023103876}. 
In this scenario, vast amounts of data are generated from multiple sensors and operational logs, necessitating effective methods to identify anomalies that can indicate equipment malfunctions, process inefficiencies, or potential safety hazards. Early detection of such anomalies is crucial for maintaining operational efficiency, ensuring predictive maintenance, and preventing costly failures. This challenge is intensified by the nature of Big Data, characterized by its high volume, velocity, and variety.

Anomaly detection (AD) is increasingly becoming a critical research challenge in industrial applications. In particular, Multivariate Time Series (MTS) data, which involves observing multiple variables over time, poses additional complexities. 
Moreover, the industrial environment usually presents issues related to data quality.
Specifically, measured MTS could present noise resulting in a more difficult AD process.
Traditional anomaly detection methods, such as Recurrent Neural Networks or reconstruction-based approaches, often struggle with multivariate time series (MTS) characterized by complex temporal dependencies and cross-variable interactions \cite{pires2019high}. 
For this reason, the need arises for a robust tool to detect anomalies within MTS, especially in industrial settings.

To address these challenges, embedding techniques \cite{CAO199875} have been proposed as a means to represent multivariate TS data in a transformed space, aiming to capture the underlying structure and relationships within the data more effectively.
Autoencoders (AE) have emerged as a powerful technique for creating embeddings of MTS data \cite{engproc2022018023}. Jointly with AEs, various ML algorithms are commonly used for AD. For example, one-class classification could be used to detect anomalies in the embedding dataset created using the AE.

In this work, trying to enhance AD accuracy, we propose a novel framework for AD in MTS in an Industry 5.0 setting, leveraging vector embeddings with integrated temporal information to improve the AD process.
Specifically, we designed a Time2Vec-inspired AE to create a vector representation of MTS, mapping temporal and spatial dependencies. Then, we perform AD on the embeddings using a variety of ML models, including One-Class Support Vector Machine, Isolation Forest, Elleptic Envelop, One-Class Support Data Distribution, and Local Outlier Factor.

We tested this framework in an industrial case provided by Bonfiglioli, a global leading manufacturer of gear motors and drive systems. The aim is to detect anomalies in the gear production process by analyzing TS data from various sensors installed on the machinery.
In this scenario, we compared our solution with a reconstruction-based method, which served as the baseline. This demonstrated that our Time2Vec-based AE either matched or, in some cases, significantly exceeded baseline performance.

The primary contributions of this paper are: 
$(i)$ we develop a Time2Vec-based AE for generating high-dimensional embeddings from multivariate TS data;
$(ii)$ we evaluate the effectiveness of this approach in detecting anomalies using various established ML techniques on the embeddings in a real industrial environment.

The paper is organized as follows: Section \ref{sec:Background} introduces background and related work on MTS AD. Section \ref{sec:scenario} describes the industrial use case at Bonfiglioli. Section \ref{sec:proposed_solution} introduces our ad-hoc Time2Vec-based autoencoder approach. Section \ref{sec:exp} presents the experimental evaluation of our model. Section \ref{sec:conclusions} concludes the paper and outlines future work.
 \section{Background \& Related Work} \label{sec:Background}
AD in TS data \cite{Schmidl2022AnomalyDI} is essential for maintaining efficiency and safety in various industrial contexts. An anomaly can be either an isolated observation or a sequence of observations that substantially differ from the normal distribution. Each TS data point is not just an individual measurement, but also part of a temporal sequence, where each value depends on those that come before it. This interconnected nature means that patterns and trends unfold over time, making it possible to utilize historical data to predict and understand future behavior. 
In analyzing large sets of MTS, it is crucial to identify interactions between features. Traditional linear autoregressive models are often insufficient, necessitating new approaches to handle non-linear and non-stationary TS interactions \cite{Tank2018DiscoveringII}. 
Creating effective AD models is very challenging due to the complexities of TS data. An important issue is the imbalance in training data, where instances of normal operations greatly outnumber rare occurrences of anomalies. This makes it hard for ML models to accurately learn about anomalous behavior, since there isn't enough anomalous data for effective training, and, the unpredictable nature of anomalies adds another layer of complexity. Unlike regular predictive modeling where events or outcomes are known and defined, anomalies can often show entirely new behaviors or patterns that haven't been observed or labeled before. This unpredictability requires robust and generalizable algorithms to detect deviations that fall outside the range of historical data used for training. Unsupervised deep learning methods operate without labeled examples of normal or anomalous states, making them ideal for scenarios where anomalies are rare or have not been previously identified \cite{unsup}.

Forecasting-based models for AD predict future values of a TS using historical and current data. Anomalies are identified by assessing the discrepancies between these predicted values and the actual observed values. Recurrent Neural Networks (RNNs) and Long-Short Term Memory Networks (LSTMs) are preferred for their efficacy in managing lengthy interrelated sequences and modeling intricated temporal dynamics, attributed to their ability to maintain memory over time. LSTM-based AD for multivariate TS data was first proposed in \cite{stacked} by stacking LSTM networks. Other applications of LSTM network architectures have been explored, such as AD-LTI \cite{adlti}. 

Reconstruction-based deep learning models for AD operate on the principle that normal patterns within the data can be learned and reconstructed accurately, whereas anomalies cannot be reconstructed well and thus will show significant deviations when the model attempts to reconstruct them. In particular, models for normal behavior are constructed by encoding subsequences of normal training data in latent spaces. In the test phase, the model cannot reconstruct anomalous subsequences since it is only trained on normal data. As a result, anomalies are detected by reconstructing a point/sliding window from
test data and comparing them to the actual values, by calculating a reconstruction error. AEs are among the most widely used reconstruction-based models, especially in the context of solving AD problems. They consist of two main components: an encoder that compresses the input data into a lower-dimensional representation, and a decoder that attempts to reconstruct the input data from this compressed form. The goal is to minimize the difference between the original input and its reconstruction, which helps to identify anomalies when the reconstruction error is notably high. Variational AEs introduce a probabilistic element, encoding inputs as distributions rather than fixed points in the latent space. Some implementations of these architectures for TS AD include CAE \cite{CHEVROT2022102652} and USAD \cite{usad}.

Hybrid methods for AD in TS data involve combining various techniques to enhance the overall performance of the detection system. These methods typically integrate classical statistical approaches, ML models, and deep learning architectures to handle the intricacies of TS data, such as trends, seasonality, and noise. They often blend a forecasting-based model with a reconstruction-based model to obtain improved TS representations. An example is CAE-M \cite{caem}, which can model generalized patterns based on normalized data by undertaking reconstruction and
prediction simultaneously. This is achieved through the combination of a convolutional AE and LSTMs.
 
All of the mentioned methods conduct unsupervised AD on MTS data. However, they rely on raw input TS for this task, meaning that AD is performed on the raw data itself rather than on a representation of the data. 
Methods like \cite{time2feat} aim to create a representation for MTS based on feature extraction and selection. There is currently limited research on AD performed on embeddings, primarily due to the complexities involved in generating and interpreting embeddings for MTS data.
Embeddings are dense vector representations that aim to capture the inherent structure and relationships within the data, potentially offering a more nuanced view of normal versus anomalous patterns compared to raw data analysis. In the context of MTS, embeddings need to not only represent the individual data points but also their sequence and temporal context.
Higher-dimensional embeddings can be particularly beneficial in MTS data due to the complexity and richness of the data involved \cite{scalableML}. Higher-dimensional embeddings can capture complex feature relationships more effectively, preserving the essential characteristics of the data that might be lost in lower-dimensional representations. However, the use of higher-dimensional embeddings also introduces significant challenges. The primary concern is the curse of dimensionality \cite{cursedim}, which arises because the vector space grows exponentially with the addition of each dimension. As the number of dimensions increases, the data distribution across the space becomes increasingly sparse. This sparsity is problematic because the data points are spread out over a large volume, and there is less likelihood that any two points are close to each other. Consequently, traditional distance metrics become less meaningful, making it harder for ML models to discern patterns or make accurate predictions.
Previous efforts have used AEs to generate general-purpose embeddings for domain-agnostic problems \cite{bertrand2024autoencoderbased}. Additionally, specific solutions like Time2Vec (T2V) \cite{kazemi2019time2vec} have been proposed to address the unique characteristics of time-dependent data. T2V is a technique employed to represent time as a feature in neural network models effectively. It encodes time into vectors that capture both linear and periodic characteristics, enabling models to recognize and utilize temporal dynamics within data. T2V is defined as:
\begin{definition}
For each time instance $\tau$, T2V of $\tau$, denoted as $T2V(\tau)$, is a vector of size $k+1$, defined as follows:
\[
T2V(\tau)[i] = 
\begin{cases} 
\omega_i \tau + \phi_i & \text{if } i = 0, \\
F(\omega_i \tau + \phi_i) & \text{if } 1 \leq i \leq k,
\end{cases}
\]
\end{definition}
where $T2V(\tau)[i]$ is the $i^{th}$ element of $T2V(\tau)$, $F$ is a periodic activation function, and $\omega$s and $\phi$s are learnable parameters.

The application of T2V as an embedding technique allows for a more refined and structured representation of TS data. Thus, T2V serves as a bridge between traditional TS analysis and advanced AD strategies that exploit the rich information contained in embeddings.

 \section{Industrial Use Case}  \label{sec:scenario}
In this work, we developed an MTS AD tool for a Bonfiglioli machine dedicated to producing gears.
Bonfiglioli\footnote{\url{https://bonfiglioli.com}} is a prominent manufacturer with over 130 years of experience in designing and producing a wide array of gear motors, drive systems, planetary gear motors, reducers, and inverters. The company is a leader in power transmission production and increasingly adopting Industry 5.0 best practices, focusing on implementing efficient and environmentally sustainable processes.

The goal was to train an AD model to identify irregularities in the gear production process. 
As a critical component in Bonfiglioli's manufacturing line, this machine requires continuous monitoring to ensure high precision and quality in gear production.
To achieve this, Bonfiglioli collected extensive TS data from various sensors installed on the machine.

Each monitored Bonfiglioli manufacturing process was recorded in a CSV file, for a total of one million rows. 
Each row represents a specific second in time and, for each second, 104 different measurements were taken.

As a first step, considering only one file at a time, we selected 6 measurements out of 104, on the recommendation of the domain experts.
Next, we checked the presence of NaN values and deleted duplicated rows and outliers. 
The "fill forward" method was employed to substitute NaN values with the value from the previous row.
Regarding outliers, it was important to remove them due to the unlabeled nature of the dataset, which may contain noise or anomalies. 
To reach this goal, we deleted rows where at least one value was below the first or above the third quantile, calculated on the single column. 
Some files contained only 90 rows, while others contained over 10,000 due to the varying duration of the manufacturing process.

To shorten the TS in the longer files we applied resampling, aggregating rows belonging to the same time window into a single value, calculating the values average, and preserving only the time index of the last row in the time window. 
The dataset was split into multivariate fixed-size TS of 100 seconds.
The examples shorter than 100 seconds were extended by applying constant padding, specifically prolonging the last value.
Pre-processing produced a final dataset composed of a total of 2950 multivariate TS each 100 seconds long, with 6 different values for each second. 
 \section{Ad-hoc Time2Vec-based Autoencoder} \label{sec:proposed_solution}
To address the challenges related to the aforementioned industrial scenario, we developed an ML model capable of extracting embeddings from MTS. 
Next, we conducted AD using various one-class classification ML algorithms.

While our approach does not compress information, it uses high-dimensional embeddings to effectively describe the multivariate features of TS data. This method aims to maintain a detailed and informative representation where both spatial and temporal attributes are preserved.

\begin{figure}[htb]
	\centering
	\includegraphics[width=0.8\columnwidth]{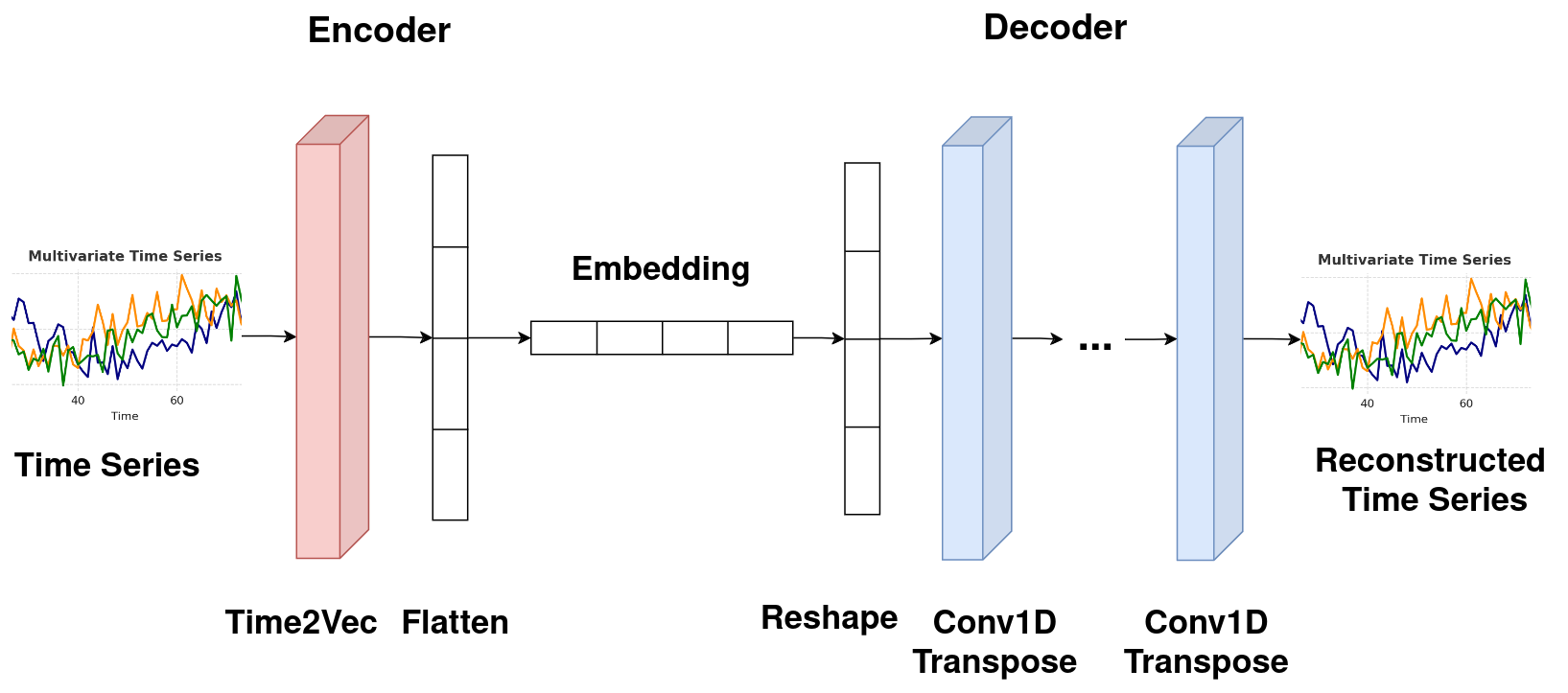}
    	\caption{\label{fig:T2V_AE} \small Ad-hoc T2V-based AE Architecture.}
\end{figure}

Furthermore, as shown in \cite{kazemi2019time2vec}, vector representations of time must capture both periodic and non-periodic patterns, being invariant to time rescaling and be simple enough to be used in different architectures.
Therefore, drawing inspiration from \cite{Pealoza2020Time2VecEO_simile}, where T2V is utilized as an embedding layer within a larger architectural framework, we have developed a novel AE designed to extract vector representations from MTS data, as illustrated in Fig. \ref{fig:T2V_AE}.
As highlighted in Section \ref{sec:Background}, T2V was originally proposed to produce a better representation of time by computing a vector using the scalar notion of time as input.
However, we can assume that our TS includes an implicit notion of time, enabling us to use the TS values rather than the time.
Given a multivariate TS $X$ of dimension N$\times$F, where N is the number of steps and $F$ is the number of features and an embedding dimension K (where K is a hyperparameter), we use the function ${T2V_{MTS}}(X)$ to create a matrix representation (that is subsequently flattened into a vector). Where each column $i$ is defined as follows:

\begin{equation} 
\label{eq:formula}
{T2V_{MTS}}(X)[i] = 
    \begin{cases}
        X\, \omega_0 + b_0                    & \text{if } i = 0        \\
        \sin\left( X\, \omega_i + b_i \right) & \text{if } 1 \leq i < K \\
    \end{cases}
\end{equation}

\noindent
where the $\sin\left( \cdot \right)$ function operates element-wise. $w_{0}$ is a vector of weights (F$\times$1) and $b_{0}$ is a vector of biases (N$\times$1). $w_i$ are matrices of weights of dimension F$\times$(K-1), $b_i$ are matrices of biases (N$\times$(K-1)). By creating the embedding in this way it is possible to capture both non-periodic patterns $(Xw_0 + b_0)$ and periodic patterns $(\sin(Xw+b))$. Also by using the sine function, we make the embedding invariant to time rescaling \cite{time2vec}.

The output of this layer is flattened to obtain a 1-D vector representing the original TS. Therefore, the Decoder is composed of a reshaping layer and a series of 1-D convolutional layers that have to reconstruct the original TS starting from the previously computed embedding.

 \section{Experimental Evaluation}
\label{sec:exp}
We conducted several experiments to assess our T2V-based AE performance, specifically accuracy in detecting anomalies paying particular attention to the noise model.
We compared the results obtained using it jointly with different ML algorithms with the reconstruction-based AE representing the baseline. 

Initially, the industrial TS dataset was split into two parts: training and test set, the latter containing 10\% of examples.
We generated four additional test sets by including modified MTS from the original test set, altered by adding noise and synthetic anomalies, as it was assumed that the original dataset did not include them.
Specifically, we altered the MTS adding two kinds of anomalies. The first one is described as a \textbf{step} function and the second one is a series of \textbf{periodic} spikes sampled from a 3-component Gaussian Mixture Model.
We made this choice assuming anomalies in industrial machinery typically appear as step and periodic disturbances within the MTS data.
Moreover, to further test and validate the AEs' robustness to noise and precision, we introduced random noise, including single-point and salt-and-pepper noise, into 10\% of the test set.  
This type of noise, representing sporadic, sharp disturbances, does not align with the expected anomaly in industrial machinery. The goal was to assess whether the AEs could effectively distinguish and ignore irrelevant noise, concentrating instead on significant disturbances relevant to machinery operation.
Injecting these anomalies and noise, we created two different test set versions—one where anomalies were injected in all 6 features and the other where only 4 features were affected.
This approach was taken because two of the features were almost flatlined, with minimal fluctuations, and we wanted to assess whether excluding these features would impact the AD tool's sensitivity to them.
Therefore, we tested T2V-based AE on four different test sets. 
In the first test set (A-6F) we injected anomalies in all 6 features, while the second contains synthetic noise (AN-6F).
The third (A-4F) and fourth (AN-4F) are the same as the first and second, but anomalies were injected only in 4 features.

\subsection{Time2Vec-based Autoencoder}
We trained the T2V-based AE from Section \ref{sec:proposed_solution} to create a vector representation of MTS.  
Specifically, it was trained to minimize the TS reconstruction error, computed using the L2 loss.
Given the extensive number of hyperparameters and their significant impact on the SE performance \cite{hyperparam}, we conducted a systematic search for the optimal hyperparameters using Optuna\footnote{https://optuna.org/}, a comprehensive hyperparameter optimization tool.
The tunable hyperparameters include the embedding size, the number of filters, the kernel size of the convolutional layer, the number of layers, the number of epochs, and the training batch size.
Therefore, the training process was repeated many times with different hyperparameters automatically optimized.
Optuna was configured to minimize the reconstruction error computed using Dynamic Time Warping (DTW) between the actual series and the reconstructed ones.
We did not employ DTW for training the AE because it is non-differentiable.

After approximately 150 training trials, we empirically noticed that the loss did not decrease any further. Consequently, we selected the AE with the lowest DTW score. 
In the latter, at the end of the training process, the K parameter in Equation \ref{eq:formula}, representing dimensionality (number of features) for each time instance in embeddings, was set to 7 and the number of convolutional layers in the decoder was set to 3.  
Following the training of the AE, we used the encoder to obtain embeddings of the training set, which has subsequently been used to perform AD using many ML techniques.
We chose different state-of-the-art one-class classification ML algorithms to detect anomalies within the embedding test datasets \cite{onclas_classification_survey,survey2}: Local Outlier Factor (LOF), Isolation Forest (IF), One-class Support Vector Machine (OCSVM), Deep Support Vector Data Description (Deep SVDD), and Elliptic Envelope (EE).

\subsection{Recontruction-based Autoencoder}
We trained another AE, whose reconstruction error was used to perform AD for performance comparison purposes. 
This choice was made to compare our solution (T2V-based AE) with a standard, well-tested method in AD literature. 
In this architecture, the Encoder and Decoder consist of a series of 1-D convolutional layers.
This AE, like the T2V-based one, was trained using Optuna to optimize the hyperparameters. During the training process, L2 loss is used while trials are evaluated using DTW. 
AD in this case is carried out by measuring the reconstruction error using a combination of L2 Loss, Mean Absolute Error (MAE), and DTW. DTW is employed to calculate the similarity between TS, accounting for differences in their lengths. MAE and L2 Loss are utilized to emphasize significant errors.

\subsection{Results}
We evaluated the performance using F1-Score, Precision, and Recall across four different test sets, with anomalies (A) and noise (N) injected, affecting either six (6F) or four (4F) features.
As shown in Table \ref{tab:results_embedding}, T2V-based AE mirrors the baseline (reconstruction-based AE) in relatively clean and predictable settings, such as the A-6F test set where it achieves an F1-score of 0.99.
EE and the baseline excel in scenarios where the test data is without noise and where simulated anomalies involve either step and periodic components on all of the six features of the MTS.
However, EE, LOF and Deep SVDD algorithms achieved slightly worse results. 

It is worth noting how reconstruction-based AE performance, specifically Precision, significantly declined with the introduction of noise or in situations where anomalies do not affect all features, showing its fine-tuning for expected anomalies but difficulty in differentiating sporadic noise from actual anomalies. This limitation suggests a tendency to misclassify such noise as significant anomalies, potentially leading to many false positives.
It is worth noting the best results in A-4F and AN-4F were achieved using Deep SVDD paired with the T2V-based AE, reaching F1-scores of 0.91 and 0.78 (against 0.73 and 0.43 of the baseline), respectively.
This suggests that, in this specific use case, the T2V-based AE with EE and Deep SVDD, especially in scenarios with noise injection, is the most effective method for anomaly detection.
Nonetheless, the T2V-based AE joint with LOF also showed stable performance across all test sets.
These results demonstrate that performing AD on T2V-based AE embedding is capable of mirroring, or even significantly exceeding the baseline. 
This underscores our solution's advantage in handling various anomaly detection scenarios, making it a versatile choice for diverse and unpredictable environments.

\begin{table}[t]
    \centering
    \renewcommand{\arraystretch}{1.3} \begin{tabular}{|m{1.3cm}<{\centering}|m{1.5cm}<{\centering}m{1.5cm}<{\centering}m{1.5cm}<{\centering}|m{1.5cm}<{\centering}m{1.5cm}<{\centering}m{1.5cm}<{\centering}|}
        \hline
        & \multicolumn{3}{c|}{\textit{Reconstruction-based AE}} & \multicolumn{3}{c|}{\textit{T2V-based AE + IF}} \\
        \hline
        Test Set & \textit{F1-Score} & \textit{Precision} & \textit{Recall} & \textit{F1-Score} & \textit{Precision} & \textit{Recall} \\ 
        \hline
        \textit{A-6F} & \textbf{0.99} & 0.99 & \textbf{0.99} & 0.41 & 0.35 & 0.51 \\ 
        \textit{AN-6F} & 0.61 & 0.44 & \textbf{0.99} & 0.34 & 0.27 & 0.46 \\ 
        \textit{A-4F} & 0.73 & \textbf{0.98} & 0.58 & 0.53 & 0.49 & 0.57 \\ 
        \textit{AN-4F} & 0.43 & 0.34 & 0.58 & 0.42 & 0.31 & 0.64 \\
        \hline  
        & \multicolumn{3}{c|}{\textit{T2V-based AE + EE}} & \multicolumn{3}{c|}{\textit{T2V-based AE + OCSVM}} \\
        \hline
        \textit{A-6F} & \textbf{0.99} & \textbf{1} & \textbf{0.99} & 0.69 & 0.74 & 0.65 \\ 
        \textit{AN-6F} & 0.62 & 0.45 & \textbf{0.99} & 0.6 & 0.56 & 0.65 \\ 
        \textit{A-4F} & 0.76 & 0.75 & 0.78 & 0.81 & 0.78 & 0.84 \\ 
        \textit{AN-4F} & 0.35 & 0.28 & 0.46 & 0.71 & 0.62 & 0.84 \\ 
        \hline 
        & \multicolumn{3}{c|}{\textit{T2V-based AE + LOF}} & \multicolumn{3}{c|}{\textit{T2V-based AE + Deep SVDD}} \\
        \hline
        \textit{A-6F} & 0.84 & 0.92 & 0.76 & 0.83 & 0.84 & 0.83 \\ 
        \textit{AN-6F} & 0.74 & 0.72 & 0.76 & \textbf{0.81} & \textbf{0.8} & 0.83 \\ 
        \textit{A-4F} & 0.75 & 0.73 & 0.77 & \textbf{0.91} & 0.85 & \textbf{0.97} \\ 
        \textit{AN-4F} & 0.78 & 0.72 & 0.85 & \textbf{0.87} & \textbf{0.79} & \textbf{0.97} \\ 
        \hline 
    \end{tabular}
    \vspace{0.2cm}
    \caption{\small Comparison of AD performance using the T2V-based AE jointly with many ML algorithms (IF, EE, OCSVM, LOF, Deep SVDD) and the reconstruction-based AE. In the test set names \textit{A} means Anomaly injected, \textit{N} means Noise injected, and -\textit{4/6}F means \textit{4/6} features are affected by anomaly injection. In bold is highlighted the highest score for each test set.}
    \vspace{-20pt}
    \label{tab:results_embedding}
\end{table} \section{Conclusion and future work}
\label{sec:conclusions}
In this paper, we presented an AD framework that leverages a T2V-based autoencoder to extract vector representation from MTS. This brings several advantages, such as the flexibility of using many ML techniques that provide consistent and high-performance metrics across varying conditions.
We demonstrated how our solution, compared with a reconstruction-based method, results in higher performance, stability, and robustness to noise in a real industrial use case.

These excellent initial results motivate us to perform further experimentation to assess the AD performance in additional use cases and with different datasets. First of all, we are going to test our solution on other real-world industrial datasets to validate the model's practical applicability, and refine our approach based on these findings.
In addition, we plan to investigate and implement alternative methods for generating embeddings that can possibly better capture the underlying patterns in MTS, enhancing the overall performance of the AD process. This could include researching different architecture or developing new techniques.
Finally, we plan to apply our MTS embedding solution beyond AD applications -- for example, to estimate accurately the Remaining Useful Lifetime (RUL) of industrial machinery.

\section{Acknowledgements}
\scriptsize
Research funded by the Italian Ministry of University and Research through PNRR, Mission 4, Component 2, Investment 1.3, Partenariato Esteso PE00000013 -- "FAIR: Future Artificial Intelligence Research" -- Spoke 8 "Pervasive AI" -- CUP J33C22002830006 (Grant Assignment Decree n. 341 adopted  15/03/2022), and Investment 1.4, Call for tender No. 1409 published on 14/9/2022, "National Centre for HPC, Big Data and Quantum Computing (HPC)" -- CUP D43C22001240001 (Grant Assignment Decree No. 1031 adopted on 17/06/2022), under NextGeneration EU programme grants.

\bibliographystyle{splncs04}
\bibliography{bibliography}

\begin{thebibliography}{10}
\providecommand{\url}[1]{\texttt{#1}}
\providecommand{\urlprefix}{URL }
\providecommand{\doi}[1]{https://doi.org/#1}

\bibitem{cursedim}
Aremu, O.O., Hyland-Wood, D., McAree, P.R.: A machine learning approach to
  circumventing the curse of dimensionality in discontinuous time series
  machine data. Reliability Engineering \& System Safety  \textbf{195},  106706
  (2020). \doi{10.1016/j.ress.2019.106706}

\bibitem{usad}
Audibert, J., Michiardi, P., Guyard, F., Marti, S., Zuluaga, M.A.: Usad:
  Unsupervised anomaly detection on multivariate time series. In: Proceedings
  of the 26th ACM SIGKDD International Conference on Knowledge Discovery \&
  Data Mining. p. 3395–3404. KDD '20, Association for Computing Machinery,
  New York, NY, USA (2020). \doi{10.1145/3394486.3403392}

\bibitem{hyperparam}
Berahmand, K., Daneshfar, F., Salehi, E., Li, Y., Xu, Y.: Autoencoders and
  their applications in machine learning: a survey. Artificial Intelligence
  Review  \textbf{57} (02 2024). \doi{10.1007/s10462-023-10662-6}

\bibitem{bertrand2024autoencoderbased}
Bertrand, J.H., Gargano, J.P., Mombaerts, L., Taws, J.: Autoencoder-based
  general purpose representation learning for customer embedding (2024)

\bibitem{time2feat}
Bonifati, A., Buono, F.D., Guerra, F., Tiano, D.: Time2feat: learning
  interpretable representations for multivariate time series clustering. Proc.
  VLDB Endow.  \textbf{16}(2),  193–201 (oct 2022).
  \doi{10.14778/3565816.3565822}

\bibitem{CAO199875}
Cao, L., Mees, A., Judd, K.: Dynamics from multivariate time series. Physica D:
  Nonlinear Phenomena  \textbf{121}(1),  75--88 (1998).
  \doi{10.1016/S0167-2789(98)00151-1}

\bibitem{CHEVROT2022102652}
Chevrot, A., Vernotte, A., Legeard, B.: Cae: Contextual auto-encoder for
  multivariate time-series anomaly detection in air transportation. Computers
  \& Security  \textbf{116},  102652 (2022). \doi{10.1016/j.cose.2022.102652}

\bibitem{colombi2024mlops}
Colombi, L., Gilli, A., Dahdal, S., Boleac, I., Tortonesi, M., Stefanelli, C.,
  Vignoli, M.: A machine learning operations platform for streamlined model
  serving in industry 5.0. In: NOMS 2024-2024 IEEE Network Operations and
  Management Symposium. pp.~1--6 (2024). \doi{10.1109/NOMS59830.2024.10575103}

\bibitem{brinacolombi}
Dahdal, S., Colombi, L., Brina, M., Gilli, A., Tortonesi, M., Vignoli, M.,
  Stefanelli, C.: An mlops framework for gan-based fault detection in
  bonfiglioli’s evo plant. Infocommunications Journal  \textbf{16}(2),  2--10
  (2024). \doi{10.36244/ICJ.2024.2.1}

\bibitem{scalableML}
Echihabi, K., Zoumpatianos, K., Palpanas, T.: Scalable machine learning on
  high-dimensional vectors: From data series to deep network embeddings. In:
  Proceedings of the 10th International Conference on Web Intelligence, Mining
  and Semantics. p. 1–6. WIMS 2020, Association for Computing Machinery, New
  York, NY, USA (2020). \doi{10.1145/3405962.3405989}

\bibitem{kazemi2019time2vec}
Kazemi, S.M., Goel, R., Eghbali, S., Ramanan, J., Sahota, J., Thakur, S., Wu,
  S., Smyth, C., Poupart, P., Brubaker, M.: Time2vec: Learning a vector
  representation of time. arXiv preprint arXiv:1907.05321  (2019)

\bibitem{time2vec}
Kazemi, S.M., Goel, R., Eghbali, S., Ramanan, J., Sahota, J., Thakur, S., Wu,
  S., Smyth, C., Poupart, P., Brubaker, M.A.: Time2vec: Learning a vector
  representation of time. CoRR  (2019), \url{http://arxiv.org/abs/1907.05321}

\bibitem{survey2}
Khan, S.S., Madden, M.G.: One-class classification: taxonomy of study and
  review of techniques. The Knowledge Engineering Review  \textbf{29}(3),
  345–374 (2014). \doi{10.1017/S026988891300043X}

\bibitem{stacked}
Malhotra, P., Vig, L., Shroff, G., Agarwal, P.: Long short term memory networks
  for anomaly detection in time series. In: ESANN 2015 proceedings, European
  Symposium on Artificial Neural Networks (04 2015)

\bibitem{Pealoza2020Time2VecEO_simile}
Pe{\~n}aloza, V.: Time2vec embedding on a seq2seq bi-directional lstm network
  for pedestrian trajectory prediction. Res. Comput. Sci.  \textbf{149},
  249--260 (2020), \url{https://api.semanticscholar.org/CorpusID:231780785}

\bibitem{pires2019high}
Pires, A.M., Branco, J.A.: High dimensionality: The latest challenge to data
  analysis (2019)

\bibitem{Schmidl2022AnomalyDI}
Schmidl, S., Wenig, P., Papenbrock, T.: Anomaly detection in time series: A
  comprehensive evaluation. Proc. VLDB Endow.  \textbf{15},  1779--1797 (2022),
  \url{https://api.semanticscholar.org/CorpusID:250331153}

\bibitem{onclas_classification_survey}
Seliya, N., Abdollah~Zadeh, A., Khoshgoftaar, T.: A literature review on
  one-class classification and its potential applications in big data. Journal
  of Big Data  \textbf{8}(1), ~122 (2021). \doi{10.1186/s40537-021-00514-x}

\bibitem{unsup}
Siegel, B.: Industrial anomaly detection: A comparison of unsupervised neural
  network architectures. IEEE Sensors Letters  \textbf{4}(8), ~1--4 (2020).
  \doi{10.1109/LSENS.2020.3007880}

\bibitem{Tank2018DiscoveringII}
Tank, A.: Discovering interactions in multivariate time series (2018),
  \url{https://api.semanticscholar.org/CorpusID:127719466}

\bibitem{engproc2022018023}
Tziolas, T., Papageorgiou, K., Theodosiou, T., Papageorgiou, E., Mastos, T.,
  Papadopoulos, A.: Autoencoders for anomaly detection in an industrial
  multivariate time series dataset. Engineering Proceedings  \textbf{18}(1)
  (2022). \doi{10.3390/engproc2022018023}

\bibitem{VENANZI2023103876}
Venanzi, R., Dahdal, S., Solimando, M., Campioni, L., Cavalucci, A., Govoni,
  M., Tortonesi, M., Foschini, L., Attana, L., Tellarini, M., Stefanelli, C.:
  Enabling adaptive analytics at the edge with the bi-rex big data platform.
  Computers in Industry  \textbf{147},  103876 (2023).
  \doi{10.1016/j.compind.2023.103876}

\bibitem{adlti}
Wu, W., He, L., Lin, W., Su, Y., Cui, Y., Maple, C., Jarvis, S.: Developing an
  unsupervised real-time anomaly detection scheme for time series with
  multi-seasonality. IEEE Transactions on Knowledge and Data Engineering
  \textbf{34}(9),  4147--4160 (2022). \doi{10.1109/TKDE.2020.3035685}

\bibitem{caem}
Zhang, Y., Chen, Y., Wang, J., Pan, Z.: Unsupervised deep anomaly detection for
  multi-sensor time-series signals. IEEE Transactions on Knowledge and Data
  Engineering  \textbf{35}(2),  2118--2132 (2023).
  \doi{10.1109/TKDE.2021.3102110}

\end{thebibliography}

\end{document}